\documentclass[10pt,twocolumn,letterpaper]{article}

\usepackage[utf8]{inputenc}
\usepackage{textcomp}
\usepackage{graphicx}
\usepackage{subcaption}
\usepackage[dvipsnames]{xcolor}

\usepackage[pagenumbers]{wacv} 

\definecolor{darkgreen}{RGB}{21, 128, 61}
\definecolor{orange}{RGB}{217, 119, 6}
\definecolor{wacvblue}{rgb}{0.21,0.49,0.74}
\usepackage[pagebackref,breaklinks,colorlinks,allcolors=wacvblue]{hyperref}
\usepackage{amssymb}
\def\wacvPaperID{*****} 
\def\confName{WACV}
\def\confYear{2027}

\title{MiD-VLM: Detecting and Grounding Global Semantic Misinformation in Image-Caption Pairs}

\author{Tarunya Prasad, Bharatesh Chakravarthi\\
Arizona State University\\
{\tt\small tprasad6@asu.edu, bharatesh@asu.edu}
}

\begin{document}
\maketitle

\begin{abstract}

Semantic misinformation can arise when an authentic image is paired with a caption that changes or distorts its narrative. Because the image itself may contain no visible signs of manipulation, methods designed to detect pixel-level artifacts can miss these misleading image-caption pairs. We study this problem using DGM4+, a dataset designed to evaluate the detection of global scene and semantic inconsistencies. We propose a multimodal model that combines image-text representations from SigLIP with visual scene features from DINOv2 through a projection MLP, and adapts Qwen2-VL-7B using LoRA. The model jointly predicts whether an image-caption pair is authentic, identifies the manipulation type, and grounds the misleading caption tokens. On the DGM4+ validation split, the model achieves an F1 score of $0.9679$ for binary authenticity classification. These results suggest that parameter-efficient adaptation of vision-language models is promising for detecting semantic inconsistencies, while also motivating further work on precise token-level grounding.

\end{abstract}

\section{Introduction}
\label{sec:intro}
Images in news and social media are often interpreted together with their captions. A genuine photograph paired with a caption describing a different event, location, or action can convey a misleading narrative even when the image itself is unaltered. In these cases, the image and caption may each appear plausible in isolation, the inconsistency emerges from their semantic relationship. Detecting such misinformation therefore requires reasoning across modalities. Methods that rely primarily on visual artifacts, such as blending boundaries or face-swap traces, may not capture these scene-level or narrative-level inconsistencies.
Prior work has investigated multimodal media manipulation through both detection and localization. The DGM$^4$ dataset and its associated HAMMER model provide a framework for identifying and grounding manipulations in image-text pairs \cite{shao2023detectinggroundingmultimodalmedia,hammer}. 

\begin{figure}[t]
  \centering
  \includegraphics[width=1\linewidth]{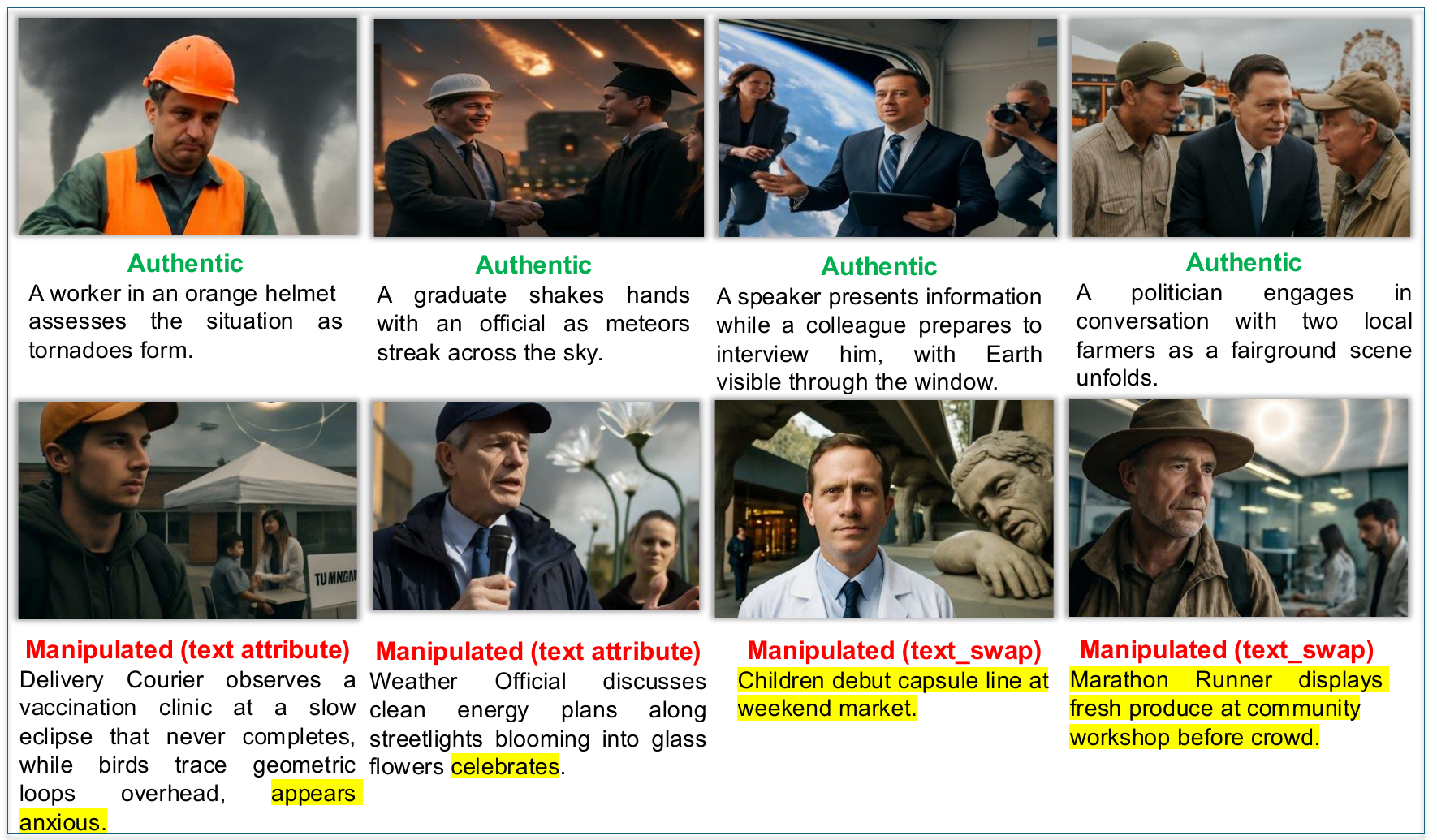}
  \caption{Examples from the DGM4+ dataset. The top row shows authentic image-caption pairs while the bottom row shows semantically manipulated pairs, with their tokens.
}
  \label{fig:teaser}
\end{figure}

However, global inconsistencies pose a distinct challenge, an image may be visually coherent while its caption describes an implausible scene or alters the meaning of the depicted event. To study this setting, DGM$^4$+ extends DGM$^4$ with $5,000$ additional image-caption pairs designed to evaluate global scene and semantic mismatches \cite{singh2025dgm4plus}. Figure \ref{fig:teaser} shows a few examples from the dataset. We focus on this extension and investigate methods for detecting these mismatches and identifying the caption content associated with them.
Given an image and a caption, our task is to predict whether the pair is authentic or manipulated, classify the manipulation type, and identify the misleading caption tokens. 
The DGM$^4$+ data includes two manipulation types, \texttt{text\_swap}, which introduces a substantially different narrative, and \texttt{text\_attribute}, which changes descriptive details while retaining the core event. These tasks require a model to assess both the overall compatibility of the image and caption and the local textual evidence that contributes to a mismatch.

We present MiD-VLM, a multimodal framework for this identification task. It combines image-text representations from SigLIP \cite{zhai2023sigmoidlosslanguageimage} with visual scene representations from DINOv2 \cite{oquab2024dinov2learningrobustvisual}. A projection MLP fuses these features, which are used alongside a LoRA-adapted Qwen2-VL-7B model for task-specific prediction \cite{wang2024qwen2vlenhancingvisionlanguagemodels}. The model has separate prediction heads for binary authenticity classification, manipulation-type prediction, and token-level caption grounding. LoRA updates approximately 0.12\% of the $8.3$ billion parameters in Qwen2-VL-7B, providing a parameter-efficient approach to adaptation.
This work makes three primary contributions.
\begin{itemize}
    \item We investigate multimodal detection of global scene and narrative inconsistencies using the DGM$^4$+ image-caption dataset.
    \item We implement a joint prediction framework for pair authenticity, manipulation type, and misleading caption-token grounding using fused SigLIP and DINOv2 features with a LoRA-adapted vision-language model.
    \item We evaluate the framework on the DGM$^4$+ validation split and analyze encoder ablations and token-grounding behavior.
\end{itemize}
On the DGM$^4$+ validation split, MiD-VLM achieves a binary authenticity-classification F1 score of $0.9679$. Our ablation results examine the contribution of the visual encoders, while the grounding analysis highlights the challenge of locating specific misleading tokens even when the model correctly classifies an image-caption pair.

\section{Related Work}
\label{sec:relate}


Multimodal misinformation can arise when authentic visual content is paired with text that changes its narrative, without visible pixel-level manipulation. This motivates methods that assess cross-modal semantic consistency alongside image editing artifacts.

\subsection{Multimodal Misinformation Detection}
Multimodal misinformation detection not only considers the authenticity of individual modalities, but also whether the semantic information conveys across modalities is mutually consistent. NewsCLIPpings \cite{luo2021newsclippingsautomaticgenerationoutofcontext} studies out of context image-caption pairs, where the image and text are individually authentic but pairing conveys misleading information. This demonstrates an important limitation of purely unimodal approaches since detecting misinformation requires reasoning about the relationship between visual and textual context.

DGM$^4$ and the HAMMER model \cite{hammer} introduce joint detection and grounding of manipulations across image and text. Rather than producing only a sample level authenticity prediction, the task also requires identifying manipulated regions or textual context, encouraging models to learn better. 
UFAFormer \cite{liu2023unifiedfrequencyassistedtransformerframework} further explores this direction by incorporating frequency domain visual information as well as cross modal features.
More recently, DGM$^4$+ \cite{singh2025dgm4plus} shifts to global scene/semantic inconsistencies. In these cases, an image may remain visually realistic while becoming inconsistent with the expressed narrative in its caption. Models like hammer when tested on the extended dataset have very poor performance and accuracies of less than $20$\% as seen in Table \ref{tab:comparison}. 

\begin{table}[t]
\centering
\caption{Reported binary accuracy and F1 on DGM$^4$~\cite{shao2023detectinggroundingmultimodalmedia} and DGM$^4$+~\cite{singh2025dgm4plus}. Our result is on the DGM$^4$+ validation split.}

\label{tab:comparison}
\resizebox{0.80\linewidth}{!}{%
\begin{tabular}{llccc}
\toprule
\textbf{Method}  & \textbf{Dataset} & \textbf{Acc (\%)} & \textbf{F1 (\%)} \\
\midrule
HAMMER~\cite{hammer}   & DGM$^4$   & 93.19 & 86.39 \\
UFAFormer~\cite{liu2023unifiedfrequencyassistedtransformerframework}        & DGM$^4$   & 86.80 & 72.02 \\
HAMMER~\cite{hammer}   & DGM$^4$+  & 19.10 & --    \\
\midrule
\textbf{MiD - VLM}                      & DGM$^4$+  & \textbf{96.13} & \textbf{96.54} \\
\bottomrule
\end{tabular}%
}
\end{table}

\subsection{Vision-Language Models}
Vision Language Models \cite{bordes2024introductionvisionlanguagemodeling}, popularly referred to as VLMs are multimodal, generative AI models, capable of understanding and processing video, image and text. Building upon large-scale vision language pretraining, recent VLMs combine visual encoders with large language models to support contextual reasoning over images and natural language. Models such as Qwen2-VL \cite{wang2024qwen2vlenhancingvisionlanguagemodels} extend these capabilities through unified multimodal representations, providing a foundation for adapting large pretrained models to specific cross modal reasoning tasks. They provide a general multipurpose backbone proving to be useful foundations for specialized cross modal reasoning without training a model from scratch. However, their large parameter counts make full fine tuning computationally expensive, particularly when the dataset is relatively small. This motivates the use of parameter efficient strategies. 

\subsection{Parameter Efficient Fine Tuning}
Adapting large pretrained models to specialized downstream tasks through full fine tuning requires updating and storing billions of model parameters, resulting in significant usage of computation, memory and time. Parameter efficient fine tuning (PEFT) methods \cite{xu2023parameterefficientfinetuningmethodspretrained} address this limitation by adapting only a small subset of parameters while leaving majority of the pretrained backbone frozen. 
LoRA \cite{hu2021loralowrankadaptationlarge} is a noteworthy PEFT approach that learns low-rank updates to frozen weights and introduces trainable low rank matrices into selected layers. It freezes the original model weights and represents task sepcific weight updates using low rank matrices, thus reducing the number of trainable parameters. In this work, we employ LoRA to adapt attention projections of Qwen2-VL while keeping most of the backbone frozen. 
Based on this, we designed and built a model to solve the problem of identification of global inconsistencies from multi media as discussed in Section \ref{sec:method}.

\section{Misinformation Detection with VLMs}
\label{sec:method}
Given an input consisting of an image-caption pair, the model is expected to output a label being either authentic or manipulated with a confidence level and if it is manipulated, the manipulation type which includes text swap or text attribute. The output is also expected to have a set of token indices $G\subseteq\{1,\dots,n\}$ marking misleading tokens.

\subsection{Problem Formulation and Dataset}

Given an image $I$ and a caption $T=(t_1,\ldots,t_n)$, the model predicts pair authenticity, manipulation type, and misleading-token indices:
\begin{equation}
(I,T) \mapsto
\left(\hat{y}_{b}, \hat{y}_{t}, \hat{G}\right),
\qquad
\hat{G} \subseteq \{1,\ldots,n\}.
\end{equation}
Here, $\hat{y}_{b}\in\{\text{authentic},\text{manipulated}\}$, while
$\hat{y}_{t}\in\{\texttt{origin},\texttt{text\_swap},
\texttt{text\_attribute}\}$. The binary head also provides a confidence score.

DGM$^4$+ contains approximately $5$K curated image-caption pairs targeting global scene and narrative inconsistencies \cite{singh2025dgm4plus}. It extends DGM$^4$ (Detecting and Grounding Multi-Modal Media Manipulation), a VisualNews-based dataset of more than $230$K samples \cite{liu2021visualnewsbenchmarkchallenges}. While DGM$^4$ focuses on local manipulations, DGM$^4$+ includes broader inconsistencies, such as implausible scene context.
We use the DGM$^4$+ extension, split into $3,500$ training, $750$ validation, and $750$ test pairs ($70$/$15$/$15$). The training set contains $1,400$ authentic, $1,050$ \texttt{text\_swap}, and $1,050$ \texttt{text\_attribute} examples. The validation set contains $254$ authentic and $446$ manipulated examples, test set labels are withheld. Each record includes an image-caption pair and annotations for the manipulation class and affected text tokens, with image region metadata where available.

\subsection{Architecture Overview}
\begin{figure}[t]
    \centering
    \begin{tabular}{cc}
        \includegraphics[width=0.80\linewidth]{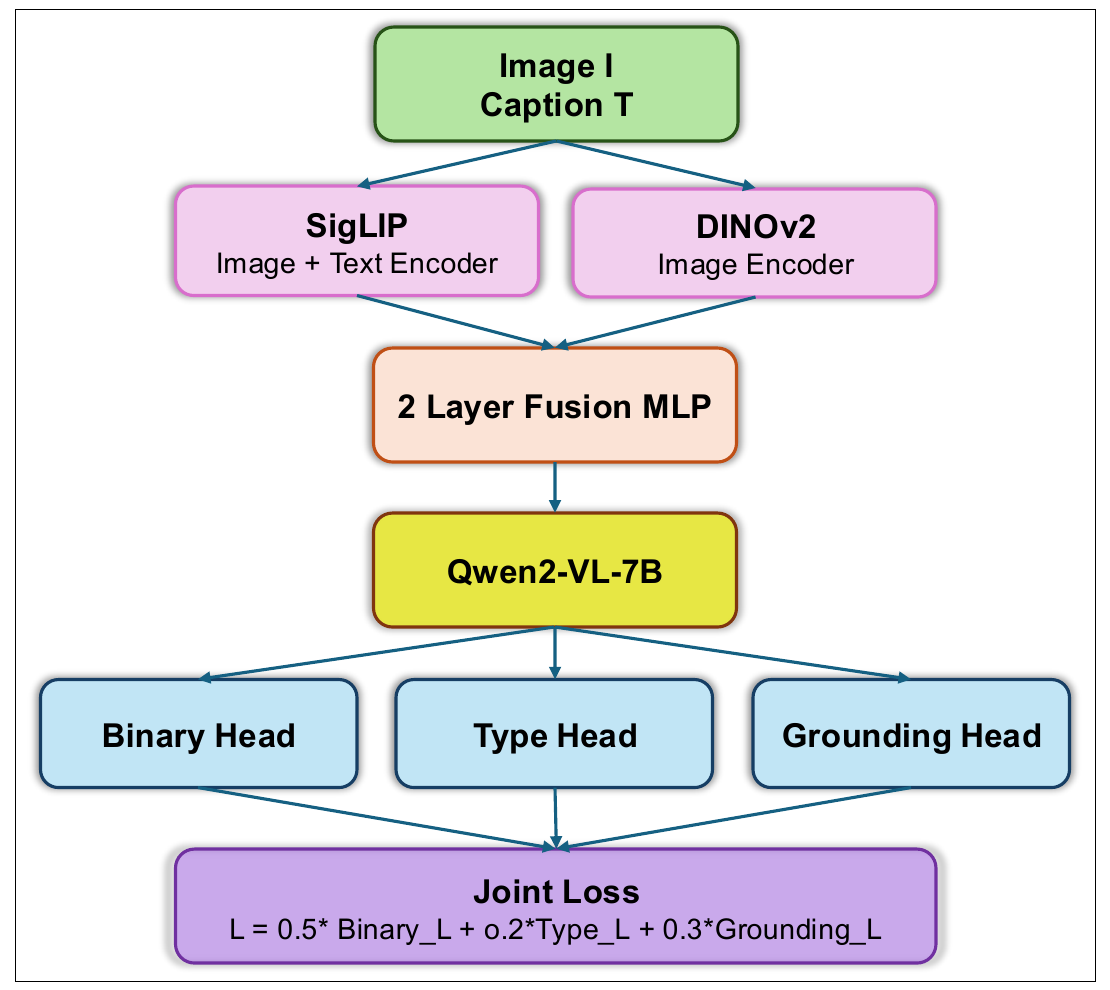} 
    \end{tabular}
    \caption{The three stage architecture of the MiD-VLM system.}
    \label{fig:arch}
\end{figure}
The proposed architecture consists of three stages: feature extraction, feature fusion and conditioning, and multitask prediction as seen in Figure \ref{fig:arch}. The image and caption pass through two frozen encoders in parallel, the outputs are concatenated and projected by a fusion multilayer perceptron (MLP) which is then injected as a learned prefix into the LoRA adapted Qwen2-VL-7B. Three shallow heads produce binary, type and grounding outputs. We will see each stage in detail below.

\subsection{Dual Encoder Feature Extraction}
The Dual encoder is used for cross modal alignment and yields image and text embeddings.

\begin{equation}
(\mathbf{s}_I, \mathbf{s}_T)
=
f_{\mathrm{SigLIP}}(I,T),
\end{equation}

\begin{equation}
\mathbf{s}_I, \mathbf{s}_T
\in \mathbb{R}^{1152}.
\end{equation}
We concatenate the SigLIP image embedding, SigLIP text embedding and DINOv2 CLS (classification) token into a single feature vector. A two layer MLP with Gaussian error linear unit (GELU) activation and layer normalization projects the features into a fused representation.  
\begin{equation}
\mathbf{d}_I
=
f_{\mathrm{DINOv2}}(I),
\end{equation}
\begin{equation}
\mathbf{d}_I
\in \mathbb{R}^{1024}.
\end{equation}
\begin{equation}
\mathbf{d}_I
=
f_{\mathrm{DINOv2}}^{\mathrm{CLS}}(I)
\in \mathbb{R}^{1024}.
\end{equation}

\begin{equation}
\mathbf{z}
=
[\mathbf{s}_I;\mathbf{s}_T;\mathbf{d}_I]
\in \mathbb{R}^{3328}.
\label{eq:feature_concat}
\end{equation}








\subsection{Multi-Task Prediction Heads}

Given the shared representation $\mathbf{h}$, three task-specific heads predict pair authenticity, manipulation type, and misleading caption tokens. The binary head outputs
\begin{equation}
\hat{\mathbf{y}}_{b}
=
\mathrm{softmax}\!\left(g_b(\mathbf{h})\right)
\in \mathbb{R}^{2},
\end{equation}
with predicted label $\hat{y}_b=\arg\max_c \hat{y}_{b,c}$ and confidence
$C=\max_c \hat{y}_{b,c}$. The manipulation-type head predicts one of
\texttt{origin}, \texttt{text\_swap}, or \texttt{text\_attribute}:
\begin{equation}
\hat{\mathbf{y}}_{t}
=
\mathrm{softmax}\!\left(g_t(\mathbf{h})\right)
\in \mathbb{R}^{3}.
\end{equation}
The grounding head assigns a score to each of 64 token positions:
\begin{equation}
\hat{\mathbf{g}}
=
\sigma\!\left(g_g(\mathbf{h})\right)
\in [0,1]^{64},
\qquad
\hat{G}
=
\{i \mid \hat{g}_i > \tau\},
\quad \tau=0.3,
\end{equation}
where $\sigma(\cdot)$ is the sigmoid function and $\hat{G}$ is the set of predicted misleading-token indices.
We train the model with a weighted sum of the binary classification, token-grounding, and manipulation-type losses. Their respective weights are $0.5$, $0.3$, and $0.2$.

\section{Experiments and Results}
\label{sec:exp}

We implemented the models in PyTorch and trained them on four NVIDIA A100 GPUs (80\,GB each) using the Sol supercomputer. Training used AdamW~\cite{loshchilov2019decoupledweightdecayregularization}, a learning rate of $2\times10^{-4}$, weight decay of $10^{-2}$, and a cosine annealing schedule. Training was capped at $100$ epochs with early stopping, which terminated after five epochs. Each GPU used a batch size of 4 with gradient accumulation over four steps.
SigLIP and DINOv2 remained frozen. We adapted Qwen2-VL-7B with LoRA ($r=16$, $\alpha_{\mathrm{LoRA}}=32$) on the query, key, value, and output projections. This trained approximately $0.12$\% of the model's $8.3$ billion parameters. Training took approximately six minutes.

\begin{figure}[t]
  \centering
  \includegraphics[width=0.85\linewidth]{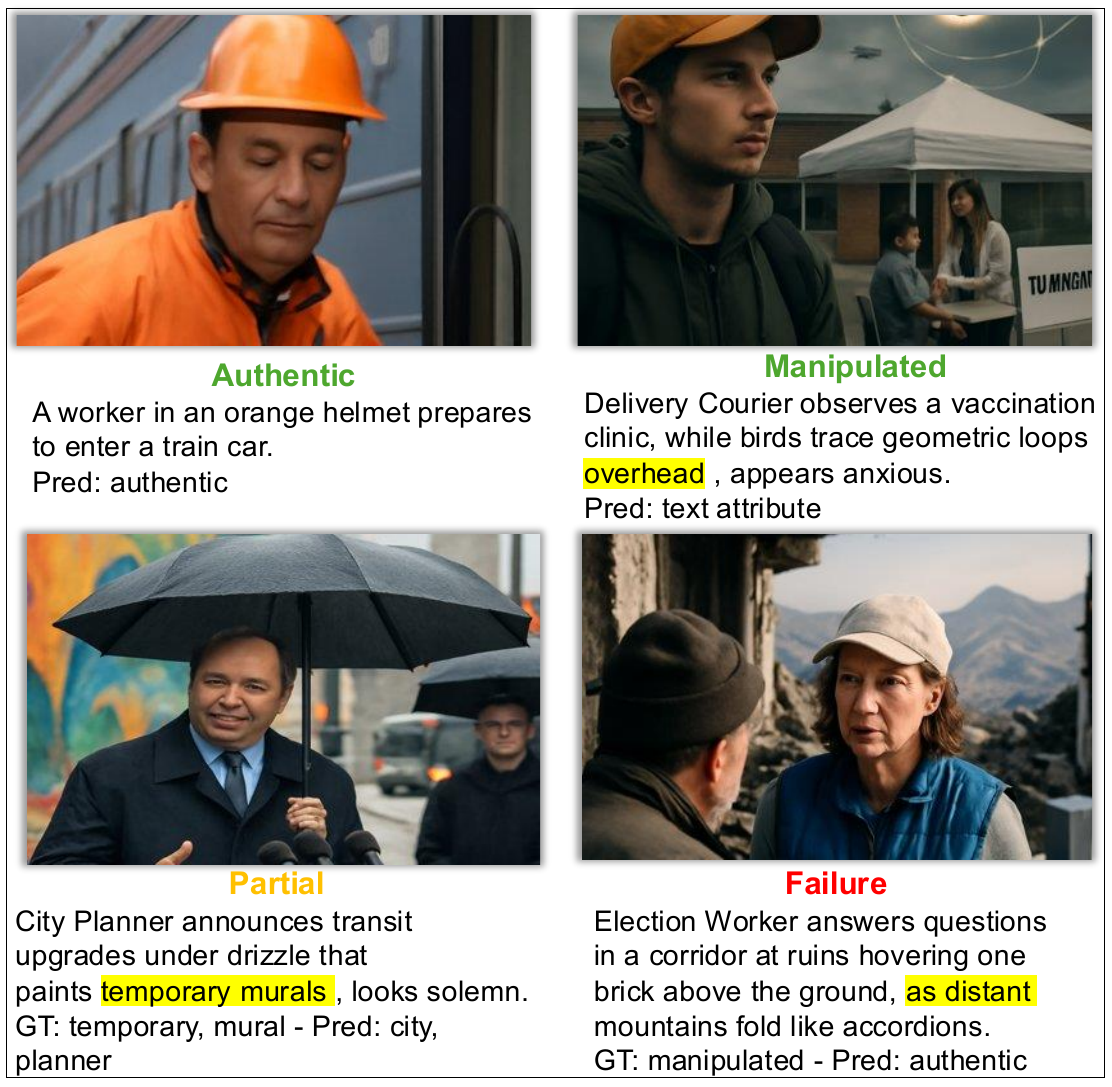}
    \caption{DGM$^4$+ validation examples: correct predictions with accurate grounding, partial grounding, and a false negative.}
  \label{fig:quali_ex}
\end{figure}

\subsection{Evaluation Metrics}
Performance was evaluated using the PP-MisDet Challenge metric, which combines binary detection, textual grounding, and manipulation-type classification:
\begin{equation}
\text{Score} =
\alpha F1_{\text{binary}}
+ \beta F1_{\text{text}}
+ \gamma F1_{\text{type}},
\label{eq:score}
\end{equation}
where $\alpha=0.5$, $\beta=0.3$, and $\gamma=0.2$.

\subsection{Validation Performance}

\begin{table}[t]
\centering
\caption{Validation performance across training epochs.}
\label{tab:validation}
\begin{tabular}{c|ccc}
\hline
\textbf{Epoch} & \textbf{Val. Loss} & \textbf{Accuracy} & \textbf{F1} \\
\hline
1 & 0.1636 & 93.20 & 0.9438 \\
2 & 0.1153 & 94.40 & 0.9541 \\
3 & 0.0975 & 95.87 & 0.9654 \\
\textbf{4} & \textbf{0.0818} & \textbf{96.13} & \textbf{0.9679} \\
5 & 0.0844 & 95.87 & 0.9660 \\
\hline
\end{tabular}
\end{table}
Figure \ref{fig:quali_ex} shows representative successful, partial and failed predictions illustrating the model's strengths in classification and limitations in textual grounding.
Performance improved through epoch 4, reaching $96.13$\% accuracy and an F1 score of $0.9679$ with the lowest validation loss of $0.0818$ (see Table~\ref{tab:validation}). Performance declined slightly at epoch $5$, and the epoch $4$ checkpoint was selected. The rapid convergence suggests that the pretrained representations require limited task-specific adaptation. 

\subsection{Encoder Ablation}
\begin{table}[ht]
\centering
\caption{Encoder ablation on the DGM$^4$+ validation set.}
\label{tab:ablation}
\resizebox{\columnwidth}{!}{%
\begin{tabular}{lcccc}
\toprule
\textbf{Encoder} &
\textbf{F1\textsubscript{binary}} &
\textbf{F1\textsubscript{text}} &
\textbf{F1\textsubscript{type}} &
\textbf{Score} \\
\midrule
DINOv2 only
& 0.9073 & 0.4192 & 0.4888 & 0.6772 \\
SigLIP only
& \textbf{0.9767} & 0.8050 & \textbf{0.9778} & 0.9254 \\
SigLIP + DINOv2
& 0.9654 & \textbf{0.8375} & 0.9711 & \textbf{0.9282} \\
\bottomrule
\end{tabular}%
}
\end{table}

We compare the full dual-encoder architecture against DINOv2-only and SigLIP-only variants in Table~\ref{tab:ablation}. SigLIP alone achieves the highest binary F1, while adding DINOv2 improves textual grounding from $0.8050$ to $0.8375$ and yields the highest overall score of $0.9282$. DINOv2 alone performs substantially worse, suggesting that SigLIP provides the primary cross-modal signal while DINOv2 contributes complementary visual information, particularly for grounding.

\subsection{Text Grounding Analysis}
Despite achieving the strongest grounding F1, the dual-encoder model produces no grounded tokens for $207$ of $446$ manipulated validation samples at $\tau=0.3$. This suggests that detecting a semantic inconsistency is easier than localizing the specific tokens responsible for it. A possible explanation is that the grounding head operates on a global fused representation, while token-level localization requires finer cross-modal correspondence.

\section{Conclusion}
\label{sec:con}
We presented a multimodal framework for semantic misinformation detection on DGM$^4$+, combining frozen SigLIP and DINOv2 representations with a LoRA-adapted Qwen2-VL backbone for joint detection, manipulation type classification, and textual grounding. The model achieved a best validation F1 of $0.9679$ while training only a small fraction of the backbone parameters. Encoder ablations further showed that DINOv2 complements SigLIP by improving textual grounding and the overall challenge score. However, difficulties in token-level grounding indicate that strong global detection does not necessarily translate to precise localization, motivating future work on finer-grained cross-modal representations.
{
    \small
    \bibliographystyle{ieeenat_fullname}
    \bibliography{main}
}

\end{document}